%% file: main_ijcai26.tex
\documentclass{article}
\usepackage{ijcai26}

\usepackage{times}
\usepackage{soul}
\usepackage{url}
\usepackage[hidelinks]{hyperref}
\usepackage[utf8]{inputenc}
\usepackage[small]{caption}
\usepackage{graphicx}
\usepackage{amsmath}
\usepackage{amsthm}
\usepackage{booktabs}
\usepackage{algorithm}
\usepackage[switch]{lineno}

\usepackage{array}
\usepackage{cleveref}
\usepackage{amsfonts}
\usepackage{nicefrac}
\usepackage{microtype}
\usepackage[T1]{fontenc}

\newtheorem{proposition}{Proposition}

\usepackage{algorithmicx}
\usepackage{amssymb}
\usepackage{subfiles}
\usepackage{mathtools}
\usepackage[english]{babel}
\usepackage[dvipsnames]{xcolor}
\usepackage{tikz, pgfplots, pgfplotstable, subcaption}
\usepackage[noend]{algpseudocode}
\usepackage{setspace}
\usepackage{yhmath}
\usepackage[font={small,stretch=0.84}]{caption}

\newcommand{\citep}[1]{\cite{#1}}
\newcommand{\citet}[1]{\citeauthor{#1} [\citeyear{#1}]}

\usepackage{cleveref}
\usepackage{mathtools}
\usepackage{stmaryrd}
\usepackage{dblfloatfix}
\crefname{section}{Section}{Sections}
\crefname{figure}{Figure}{Figures}
\crefname{mythm}{Theorem}{Theorems}
\crefname{mylem}{Lemma}{Lemmas}
\crefname{myrem}{Remark}{Remarks}
\crefname{myprop}{Proposition}{Propositions}
\crefname{equation}{Equation}{Equations}
\crefname{algorithm}{Algorithm}{Algorithms}
\crefname{table}{Table}{Tables}

\usetikzlibrary{arrows.meta, calc, topaths, positioning, automata}
\pgfplotsset{compat=1.16}
\def\addlegendimage{\csname pgfplots@addlegendimage\endcsname}
\def\BibTeX{{\rm B\kern-.05em{\sc i\kern-.025em b}\kern-.08em
    T\kern-.1667em\lower.7ex\hbox{E}\kern-.125emX}}

\algrenewcommand{\algorithmiccomment}[1]{\hfill// #1}

\newcommand{\norm}[1]{ \|   #1  \| }
\newcommand{\ind}[1]{ \mathbf{I}_{\{#1 \}} }
\newcommand{\argmax}[0]{ \mathop{\arg\max}  }
\newcommand{\argmin}[0]{ \mathop{\arg\min}  }

\newtheorem{mythm}{\bf{Theorem}}[section]

\newtheorem{myprop}{\bf{Proposition}}[section]

\newtheorem{myasp}{\bf{Assumption}}

\usepackage{multirow}

\title{Learning to Price and Stock Under Contextual and Censored Demand}

\author{
Zean Han$^1$\and
Zezhen Ding$^1$\thanks{Corresponding author.}\and
Jiheng Zhang$^1$\\
\affiliations
$^1$The Hong Kong University of Science and Technology\\
\emails
zhanax@connect.ust.hk, zdingah@connect.ust.hk, jiheng@ust.hk
}

\begin{document}

\maketitle

\input{sessions/abstract}

\input{sessions/introduction}

\input{sessions/basic_setting}

\input{sessions/algorithm_design}

\input{sessions/regret_analysis}

\input{sessions/numerical_experiments}

\input{sessions/conclusion}

\section*{Acknowledgments}

This work was supported by the Theme-based Research Project T32-615/24-R and the General Research Fund 16500023 from the Hong Kong Research Grants Council.

\bibliographystyle{named}
\bibliography{ref}

\end{document}

%% file: sessions/abstract.tex
\begin{abstract}
To make optimal joint pricing and inventory control decisions is a critical challenge for modern retailers. 
In practice, retailers face changing market conditions where demands are influenced by various contextual factors, while simultaneously dealing with the difficulty of lost sales that obscure true demand information. 
However, existing approaches often fail to account for both contextual information and censored demand observations. 
We address this gap by presenting a framework where we model demand as a linear combination of basis functions with unknown coefficients, allowing for adaptive pricing and inventory decisions that respond to changing contexts.
We propose an efficient algorithm to achieve regret bound $\mathcal{O}(K\sqrt{T}\log T)$ under concave revenue conditions and $\mathcal{O}(K^{2/3}T^{2/3}(\log T)^{1/2})$ for the general case, with matching lower bounds confirming optimality.
Extensive numerical experiments across diverse scenarios demonstrate our algorithm's effectiveness.
\end{abstract}

%% file: sessions/introduction.tex
\section{Introduction}

In recent years, joint pricing and inventory control has gained significant attention from both academia and industry. Leading retailers increasingly leverage data science to drive innovation in pricing strategies, inventory management, and supply chain optimization. This growing interest is reflected in comprehensive surveys \citep{d9be340baca34d00a7c0af41b0b75b56,Yano2004,10.1287/opre.47.2.183}. However, a critical gap remains: most existing models overlook the role of censored demand and contextual information, which are essential for adaptive decision-making. In this paper, we address this gap by studying the optimal joint pricing and inventory control strategy with lost sales that incorporates contextual information, aiming to bring theory closer to real-world applications.

In this paper, we study a joint pricing and inventory control problem with contextual information and lost sales over a finite horizon of \( T \) periods. 
At the start of each period, the seller observes contextual information that influences customer demand.
The demand function is modeled as a linear combination of \( K \) basis functions with unknown coefficients, plus an independent random noise term. 
Using the observed context and an estimated demand function, the seller simultaneously sets the selling price and inventory level, with replenishment taking effect immediately. 
After making the decisions, only the censored demand, i.e., the minimum between actual demand and available inventory, is observed as unmet demand are lost.
Leftover inventory carries over to the next period, incurring a linear holding cost, while unmet demand results in a linear lost penalty cost. 
The objective is to maximize profit, defined as sales revenue minus inventory holding and lost sales costs, across the \( T \)-period horizon.

Our problem introduces several interconnected challenges that complicate both theoretical analysis and practical implementation. 
First, the presence of censored demand obstructs direct observation of true demand signals, necessitating the development of unbiased estimators for the demand function while mitigating inherent biases from incomplete data. 
Second, the inclusion of contextual factors result in the problem being dynamic, thus traditional optimization methods which lead to static policies become inapplicable.
Both the pricing and inventory decisions must adapt to contextual information in order to achieve optimality. 
Finally, when applying our methodology to models beyond the combination of the basis functions, approximation errors arise due to structural mismatches between idealized model and real-world dynamics, demanding rigorous error quantification and robustness guarantees.
Together, these challenges necessitate a novel integration of censored data correction, context-aware online optimization, stochastic inventory control, and error-bounded approximation techniques.

\begin{table*}[ht]
    \centering
    \small
    \begin{tabular}{lcccccp{4.2cm}}
    \toprule
    \multirow{2}{*}{\textbf{Paper}} & \multirow{2}{*}{\textbf{Context}} & \textbf{Censored} & \textbf{Upper} & \textbf{Lower} & \multirow{2}{*}{\textbf{Assumptions}} \\
    & & \textbf{Demand} & \textbf{Bound} & \textbf{Bound} & \\
    \midrule
    \citep{joint_linear}        & No  & No  & $\sqrt{T}$            & N/A           & --- \\
    \citep{joint_nonparametric} & No  & Yes & $T^{\frac{1}{2}+\nu}$ & N/A           & $\sqrt{\log T}$-differentiable, concave \\
    \citep{joint_opt}           & No  & Yes & $\sqrt{T}$            & N/A           & Twice-differentiable, concave \\
    \citep{joint_opt}           & No & Yes & $T^{\frac{3}{5}}$     & $T^{\frac{3}{5}}$ & Twice-differentiable \\
    \textbf{Ours}               & Yes & Yes & $T^{\frac{2}{3}}$     & $T^{\frac{2}{3}}$ & Differentiable \\
    \textbf{Ours}               & Yes & Yes & $\sqrt{T}$            & N/A           & Twice-differentiable, concave \\
    \bottomrule
    \end{tabular}
    
        \vspace{0.5em}
        \begin{minipage}[t]{0.85\textwidth}
        \centering
        \footnotesize
        \textbf{Note:} The regrets listed above omit constant factors and $\mathrm{poly}(\log T)$ terms. Here, $\nu = \frac{1}{\sqrt{3 \ln T}} + \frac{0.25}{\sqrt{\ln T}}$.
        \end{minipage}
    \caption{Comparison of Regret Bounds under Joint Pricing and Inventory Control Problems}
    \label{tb: regret-comparison}
\end{table*}

Our contributions can be summarized as followings:
\begin{enumerate}
    \item \textbf{Context-Aware Pricing and Inventory Optimization.} 
    We propose a framework for joint pricing and inventory management (see \cref{tb: regret-comparison}) where decisions adapt to contextual information, with both pricing and inventory policies responding to changing contexts.
    
    \item \textbf{Algorithmic Framework and Regret Analysis.} 
    We propose an easy-to-implement algorithm addressing censored demand, dynamic contexts, and stochastic inventory dynamics. Under concave revenue, our method achieves $\mathcal{O}(K\sqrt{T}\log T)$ regret, matching the lower bound. Without concavity, we derive $\mathcal{O}(K^{2/3}T^{2/3}(\log T)^{1/2})$ regret and prove a lower bound $\Omega(T^{(m+1)/(2m+1)})$ in \cref{thm: lower bound} for $m$-th differentiable revenue functions, establishing minimax optimality (see \cref{tb: regret-comparison}).

    \item \textbf{Inventory Feasibility.} 
    To address infeasible target inventory levels from contextual information, we develop a queueing-theoretic method using stochastic recursion to bound regret from inventory mismatches, ensuring stable performance under stochasticity.
 
\end{enumerate}

\subsection{Related Work}

\textbf{Dynamic pricing.}
Initial research focused on non-contextual dynamic pricing \citep{non_DP_linear,DP_finiteV}. \citet{Multimodal_DP} achieved $\tilde{\mathcal{O}}(T^{(m+1)/(2m+1)})$ regret for $m$-th smooth demand functions with matching lower bounds. \citet{PlinearDP} extended this to additive linear contextual effects, establishing instance-dependent bounds. Other models assume Bernoulli purchase decisions \citep{DP_parametricF,DP_ambiguity_set,DP_cox,LP_LV,d_free_DP,Explore_UCB,DP_Fm}. Related studies \citep{d_free_DP,Explore_UCB,DP_Fm} share similar settings but differ in noise distribution assumptions, with regret bounds ranging from $\tilde{\mathcal{O}}(d_0^2 T^{2/3})$ to $\tilde{\mathcal{O}}((d_0T)^{(2m+1)/(4m-1)})$ \citep{Explore_UCB,DP_Fm}. \citet{gong2024minimax} achieved $\tilde{\mathcal{O}}(d_0^{1/3} T^{2/3})$ with minimax optimality.

\textbf{Inventory control.}
Classical inventory models, such as the newsvendor model \citep{newsvendor}, assume i.i.d. demand and use SGD-based methods for optimization, extended to multi-product settings, delayed replenishment \citep{ic_leadtime}, and quantity constraints \citep{ic_multiproduct}. \citet{minibatchIC} addressed SGD infeasibility using mini-batch strategies. Recent work incorporates demand characteristics \citep{FeatureBasedIC}, while SAA methods construct empirical distributions from demand samples \citep{SAA_newsvendor}.

\textbf{Joint pricing and inventory control.}
\citet{joint_full} first introduced joint pricing and inventory control with full information. Subsequent work \citep{joint_general} used dynamic programming, while \citet{joint_offline} proposed approximations with sample complexity analysis. In online learning settings, \citet{joint_linear} considered data-driven models without historical data. \citet{joint_nonparametric} addressed censored demand using spline interpolation and sample average methods. \citet{joint_opt} proposed ternary/binary search methods for convex/non-convex demand functions, establishing theoretical lower bounds. However, existing models ignore contextual information; once contexts are considered, optimal prices and inventory change over time, making prior algorithms inapplicable. We summarize comparisons in \cref{tb: regret-comparison}.

\textbf{Contextual bandit.}
Our policy connects to bandit algorithms \citep{banditBook,CBwithRegressionOracles,improvedUCB,misspecified_lin,supLinUCB,fasterCB}, including linear \cite{improvedUCB} and generalized linear bandits \cite{GLbandit,linContextual}. Recent work \cite{CBwithOracle,CBwithPredictableRewards,CBwithRegressionOracles} achieved optimal regret bounds for general function classes with regression oracles.

\textbf{Notations.} 
Throughout the paper, we use the following notations. For any positive integer $n$, we denote the set $\{1,2,\cdots,n\}$ as $[n]$. The cardinality of a set $A$ is denoted by $|A|$.
We use $\ind{E}$ to represent the indicator function of the event $E$. Specifically, $\ind{E}$ takes the value $1$ if $E$ happens, and $0$ otherwise.
For norms, we utilize the notation $\norm{\cdot}_p$ where $1\leq p\leq \infty$ to denote the $\ell_p$ norm.
Throughout the analysis, the notation $\tilde{\mathcal{O}}$ is employed to hide the dependence on absolute constants and logarithmic terms. It allows us to focus on the dominant behavior of the quantities involved.

%% file: sessions/basic_setting.tex
\section{Basic Setting}

Consider a company selling a single type of products over a time period of $T$ rounds. At the beginning of each round $t$, the firm observes a context $x_t$ drawn i.i.d. from an unknown distribution in a compact domain $\mathcal{X}$. Using this context, the company determines a pricing $p_t$ and inventory order-up-to decision $y_t$. The demand is modeled as $D_t = \lambda(x_t,p_t) + \epsilon_t$, where $\lambda(\cdot,\cdot)$ is a deterministic function capturing the demand-context-price curve, and $\epsilon_t$ is an i.i.d. noise random variable with mean $0$.
The distribution function and density function of $\epsilon_t$ are denoted as $F$ and $f$, respectively. The company has no prior knowledge of the function $\lambda(\cdot,\cdot)$ or the distribution $F$, and it must make its pricing and inventory decisions sequentially, relying solely on historical data to maximize the $T$ period total profit.

\textbf{Decision dynamics.}
Let $I_t$ denote the inventory level before replenishment at the start of the round $t$. An admissible policy is represented by $ \{(p_t, y_t), t \geq 1\} $, where $ p_t \in [p_{min}, p_{max}] $ is the price, and $y_t\geq I_t$
is the order-up-to inventory level.
The decision $ (p_t, y_t) $ depends only on the observable contexts and historical data prior to the round $t$.

Given any admissible policy \( \{(p_t, y_t), t \geq 1\} \), the sequence of events for each round $t$ is described as follows:

\begin{enumerate}
    \item At the beginning of each round \( t \), the firm observes the context $x_t \in \mathcal{X}\subseteq\mathbb{R}^d$ and the starting inventory level \( I_t \).
    \item The firm decides to place an order to bring the inventory level up to \( y_t \geq I_t \), and also sets the selling price \( p_t \in [p_{min}, p_{max}] \). We assume that newly ordered items arrive instantly, meaning that the ordering lead time is zero.
    \item The demand \( D_t \) realizes and is satisfied up to the available inventory.
    Any unsatisfied demand is lost and unobservable.
    Therefore, the lost-sales quantity \( (D_t - y_t)^+ \) is not observable, and the firm only observes the sales quantity (censored demand) \( o_t = \min\{D_t, y_t\} \), rather than the full realized \( D_t \).

\item At the end of round \( t \), the firm incurs a profit of
\begin{align}
q_t &= p_t \min\{D_t, y_t\} - b(D_t - y_t)^+ - h(y_t - D_t)^+ \nonumber \\
&= p_t D_t - (b + p_t)(D_t - y_t)^+ - h(y_t - D_t)^+.
\end{align}
where \( h \) and \( b \) are the per-unit holding and lost-sales penalty costs, respectively.

\item Products may be perishable, with an unknown perishability rate $\rho \in [0,1]$, which impacts the inventory carried over to the next period. The state transition is given by
\[
I_{t+1} = \max\{\rho ( {y}_t - D_t), 0\}.
\]

\end{enumerate}

\textbf{Full information benchmark.}
The firm's goal is to maximize the \( T \)-period expected total profit
$ \sum_{t=1}^{T} \mathbb{E}[q_t]$,
by constructing an admissible policy \( \{(p_t, y_t), t \geq 1\} \). This is a fundamental model for joint pricing and inventory control, with potential applications in various realistic retail scenarios.

Assuming the demand curve \( \lambda(\cdot,\cdot) \) and noise distribution \( F \) are known, we define the myopic pricing and replenishment policy as a benchmark. This policy is represented as
\[
(p^*_t, y^*_t)  = \argmax_{p, y} Q(x_t, p, y),
\]
for \( t = 1, \dots, T \), where
\begin{align*}
Q(x, p, y) = & p \mathbb{E}_{\epsilon} \left[ \min \left\{ \lambda(x,p) + \epsilon, y \right\} \right] \nonumber \\
&\quad - b \mathbb{E}_{\epsilon} \left[ (\lambda(x,p) + \epsilon - y)^+ \right] \nonumber \\
&\quad - h \mathbb{E}_{\epsilon} \left[ (y - \lambda(x,p) - \epsilon)^+ \right] \\
= & p \lambda(x,p)  - (b + p)\mathbb{E}_{\epsilon} \left[ (\lambda(x,p) + \epsilon - y)^+ \right] \nonumber \\
&\quad - h \mathbb{E}_{\epsilon} \left[ (y - \lambda(x,p) - \epsilon)^+ \right].
\end{align*}

The regret at time $t$ is defined as the loss in reward resulting from setting the price $p_t$ and the inventory level $y_t$ compared to the optimal price and inventory level.
The cumulative regret over the horizon of $T$ periods, denoted as $\mathcal{R}(T)$, is given by the expression:
$$
\mathcal{R}(T) = \sum_{t=1}^T \left[ Q(x_t,p_t^*,y_t^*) - Q(x_t,p_t,y_t)   \right].
$$
To assess performance, we consider the expected cumulative regret $\mathbb{E}[\mathcal{R}(T)]$, which accounts for the randomness in covariates and the potential randomness in the pricing policy.

\textbf{Assumptions.}
We make the following assumptions, most of which are rather standard in the literature.

First, we consider the linear additive model as the demand function,
which means that $\lambda(x,p)$ is a linear combination of the basis function.

\begin{myasp}
\label{asp: linear combination of demand function}
   There exists a set of basis functions $\{\lambda_i(x,p)\}_{i=1}^K$ and a unique unknown parameter $\theta^* = (\theta_1^*,\theta_2^*,\cdots,\theta_K^*)^\top$ such that
    $
    \lambda(x,p) = \sum_{i=1}^K \theta_i^*\lambda_i(x,p).
    $
\end{myasp}

This assumption includes many commonly used demand functions, such as \citep{DP_linear}. Next, we introduce some basic assumptions on the revenue function.

\begin{myasp}
\label{asp: differentiability of Q}
    The revenue function $Q(x,p,y)$ is continuously differentiable with respect to $x$, $p$ and $y$.
\end{myasp}

The differentiability of $Q$ is a common assumption in the literature, as seen in works such as \citep{joint_linear,joint_offline,joint_opt}. For the dynamic pricing component, this assumption is particularly relevant, as it encompasses many practical demand functions, including those discussed in \citep{DP_linear,Multimodal_DP,DP_GL}. In inventory control, several papers, such as \citep{ic_multiproduct,FeatureBasedIC}, adopt the newsvendor objective function, which is differentiable with respect to the inventory level $y$. These properties are crucial for deriving closed-form solutions for the optimal inventory level $y^*(x,p)$ given the context $x$.

\begin{proposition}
\label{prop: optimal inventory level}
    The optimal inventory level enjoys the following structure:
    $y^*(x,p) = \lambda(x,p) + z^*(p)$, where $z^*(p) = F^{-1} \left(\frac{b+p}{b+p+h}\right)$.
\end{proposition}

A context-free version of this result has appeared in \citep{FeatureBasedIC,joint_nonparametric,joint_opt}. To extend this to contextual dynamic pricing, we follow the process outlined below.
To obtain the optimal inventory level. We begin by noting that
$
\frac{\partial Q}{\partial y}  = b+p - (b+p+h) F(y-\lambda(x,p)).
$
The optimal order-up-to level thus is given by
$
y^*(x,p) = \lambda(x,p) + z^*(p),
$
where $z^*(p)$ satisfies
$
z^*(p) = F^{-1}\left(\frac{b+p}{b+p+h}\right).
$
Finally, we assume some basic properties of the distribution noise, which is rather standard in the literature.

\begin{myasp}
\label{asp: lower bound of the cdf}
     The random error \( \epsilon_t \) is bounded on the interval $[\underline{\epsilon}, \overline{\epsilon}]$, and there exist positive constants \( \kappa_1 \) and \( \kappa_2 \) such that for any \( p \in [p_{min}, p_{max}]  \) and \( \epsilon_0 \in \left[ F^{-1} \left( \frac{b + p}{b + p + h} \right) - \kappa_1, F^{-1} \left( \frac{b + p}{b + p + h} \right) + \kappa_1 \right] \), we have $f(\epsilon_0) \geq  \kappa_2.$
\end{myasp}

The first assumption on $\epsilon_t$ ensures the boundedness of the demand, which is a mild and common assumption in real-world applications \citep{joint_opt}. It is important to note that we do not require the decision maker to know the exact bounds of the noise or the demand function.
The second assumption is satisfied by many continuous random variables, including uniform and truncated Gaussian distributions.
And the third assumption needs the demand to be larger than a certain range of the noise ensuring robustness.
In the following sections, we propose our algorithm and further discuss the approximation error of the demand function.

%% file: sessions/algorithm_design.tex
\section{Algorithm Design}
\label{sec: algorithm design}

We present our algorithmic design and summarize the key challenges in this setting.

The first challenge is \emph{censored demand}: lost sales are unobservable, so we cannot directly estimate the true demand parameter~$\theta^*$. To mitigate this, we set the inventory level to scale with $\log T$, improving demand observability without knowing the true demand upper bound.

The second challenge is the \emph{period-varying optimal inventory level} due to changing contexts and perishable inventory constraints, which may make the ideal up-to-order level infeasible. We address this using a queue-based approach to control deviations from the recommended target inventory.

Recall from \cref{prop: optimal inventory level} that the optimal inventory level is $y^*(x,p) = \lambda(x,p) + z^*(p)$, where $z^*(p)$ is the safety stock level. The corresponding revenue function is
\[
G(x, p) = \max_y Q(x, p, y),
\]
where $x$ is the context and $p$ is the price. The decision-maker estimates $G(x, p)$ to determine the optimal price and then computes the corresponding target inventory.

\begin{algorithm}[h]
	\renewcommand{\algorithmicrequire}{\textbf{Input:}}
	\renewcommand{\algorithmicensure}{\textbf{Output:}}
    \algnewcommand{\LeftComment}[1]{\Statex \quad\quad  \(//\) \textit{#1}}

    \caption{Joint Pricing and Inventory Control Algorithm with Censored Demand}
    \label{alg: joint}
	\begin{algorithmic}[1]
      \Require the time horizon $T$, the length of the exploration phase $T_0$, cost parameter $h$ and $b$, basis function $\phi(x,p) = (\lambda_1(x,p),\lambda_2(x,p),\cdots,\lambda_K(x,p))^{\top}$
      \For{ round $t = 1,2,3,\cdots,T_0$ }
         \State Observe a context $x_{t}$ and current inventory level $I_t$
         \State Uniformly choose $p_t\in [p_{min},p_{max}]$ 
         \State Set the inventory level $y_t = \log T$
         \State Observe the sale $o_t = \min\{ D_t, y_t \}$
      \EndFor
    \State Estimate the parameter $\theta^*$ by 
    $$
    \hat{\theta} =  \argmin_{\theta} \sum_{t=1}^{T_0} ( o_t  - \theta^\top \phi(x_t,p_t)  )^2 
    $$
    \State Collect the samples for residuals $\eta_t = o_t  - \hat{\theta}^\top \phi(x_t,p_t) $ with $t\in [T_0]$
    \State Estimate the distribution $F$ as 
    $$
    \hat{F}(\epsilon) = \frac{1}{T_0} \sum_{t=1}^{T_0} \ind{\eta_t \leq \epsilon}
    $$
    \State Estimate the quantity $z(p)$ as 
    $$
    \hat{z}(p) = \inf\left\{u: \hat{F}(u) \geq \frac{b+p}{b+p+h} \right\}
    $$
    \State Construct the function $\hat{G}(x,p)$ using:
    \begin{align*}
    & \hat{G}(x,p) = p \hat{\theta}^\top \phi(x,p) \\
    & - \frac{b+p}{T_0} \sum_{t=1}^{T_0}[\eta_t - \hat{z}(p)]^+ - \frac{h}{T_0} \sum_{t=1}^{T_0}[\hat{z}(p) - \eta_t]^+
    \end{align*}
    \For{ round $t = T_0+1,T_0+2,\cdots, T $ }
         \State Observe a context $x_{t}$ and current inventory level $I_t$
         \State Set $p_t\in \argmax \hat{G}(x_t,p) $ 
         \State Set the inventory level $y_t = (\hat{\theta}^\top \phi(x_t,p_t) +\hat{z}({p_t}) )\vee I_t$
         \State Observe the sale $o_t = \min\{ D_t, y_t \}$
    \EndFor
   \end{algorithmic}  
\end{algorithm}

Our method splits the time horizon into two phases: exploration and commitment.
During exploration, we sample prices uniformly at random and set the inventory level high enough to fully observe demand, avoiding censoring and ensuring reliable feedback. The inventory level is tuned to balance learning accuracy and regret.
We estimate the demand function using least-squares, leveraging the linear structure of the basis functions. Least-squares is chosen because its estimation error is negligible compared to the overall regret.
Next, we estimate the total cost (including lost-sales and holding costs) by accurately estimating the optimal inventory function $z(p)$ and the residual error. We approximate the demand distribution empirically and compute $z(p)$ via the inverse CDF. With these estimates, we construct an empirical revenue function.
Overall, our algorithm is simple, efficient, and practical for joint pricing and inventory control under censored demand and perishable constraints.

%% file: sessions/regret_analysis.tex
\section{Regret Analysis}

The main result of this section is that the proposed algorithm converges to the full information benchmark at a rate of $\mathcal{O}(K\sqrt{T}\log T)$ under the strongly concave assumption, and at a rate of $\mathcal{O}(K^{2/3}T^{2/3}(\log T)^{1/2})$ without this assumption. We provide a detailed regret analysis in the following two subsections.

\subsection{Strongly Concave Revenue Function}

In this subsection, we present several additional assumptions that are standard in the literature (e.g., \cite{joint_nonparametric}, \cite{joint_opt}, \cite{joint_general}). These assumptions are crucial for establishing the theoretical framework and ensuring the robustness of our model.

First, we assume that the full information optimal price \( p_t^* \) lies in the interior of the price interval \([p_{min}, p_{max}]\) for any time \( t \). This assumption is necessary to ensure that optimal pricing decisions can be made without hitting the boundaries of the price range, which could limit the effectiveness of pricing strategies.

\begin{myasp}
\label{asp: interior point}
    The full information optimal price \( p_t^* \) lies in the interior of \( [p_{min}, p_{max}] \) for any \( t \).
\end{myasp}

Building on this, we further assume that the function \( G(x,p) \), which represents the relationship between the state variable \( x \) and price \( p \), is strongly concave in \( p \) for any given \( x \). This concavity condition is crucial, as it guarantees that the revenue generated from pricing is maximized, leading to unique optimal pricing strategies. Specifically, there exists a parameter \( \kappa_3 > 0 \) such that the second derivative of \( G \) with respect to \( p \) is always less than or equal to $-\kappa_3$.

\begin{myasp}
\label{asp: concavity}
    \( G(x,p) \) is strongly concave in \( p \) for any \( x \). In other words, there exists a parameter \( \kappa_3 > 0 \) such that
    $
    \partial^2_p G(x,p) \leq -\kappa_3
    $
    for any \( x \).
\end{myasp}

Additionally, we require that the revenue function \( Q(x,p,y) \) is twice continuously differentiable with respect to the variables \( x \), \( p \), and \( y \). This differentiability condition facilitates the application of optimization techniques and allows for smooth adjustments in response to changes in the underlying variables, ensuring a well-behaved revenue landscape.

\begin{myasp}
\label{asp:twice differentiable}
    The revenue function \( Q(x,p,y) \) is twice continuously differentiable with respect to  \( p \), and \( y \) and continuous with respect to \(x\).
\end{myasp}

Together, these assumptions create a robust foundation for analyzing the optimal pricing strategies within our model, ensuring both theoretical validity and practical applicability.

Now we are ready to give a sketch of proof of the \cref{theorem : main regret under strongly concave assumption}: under strongly concave assumption. To give an upper bound on the regret, we first decompose the regret into the following terms:
\begin{align}
        \mathcal{R}(T)
        & \leq\mathbb{E}\left[\sum_{t=1}^{T_0}(Q(x_t,p_t^*,y_t^*)-Q(x_t,p_t,y_t))\right] \nonumber \\
        &\quad +\mathbb{E}\left[\sum_{t=T_0+1}^T(Q(x_t,p_t^*,y_t^*)-Q(x_t,{p}_t,\hat{y}_t))\right] \nonumber \\
        &\quad +\mathbb{E}\left[\sum_{t=T_0+1}^T|Q(x_t,{p}_t,\hat{y}_t)-Q(x_t,p_t,y_t)|\right]. \label{decomposition of the regret}
\end{align}

where $\hat{y}_t = \hat{\theta}^T\phi(x_t, p_t) + \hat{z}(p_t)$ is the recommended inventory level and $y_t$ is the truly implemented inventory level.

Now we consider two separate situations regarding the demand model: the first is the linear additive model, while the second is the linear additive model with an approximation bias. We provide upper bounds for the four terms respectively under each demand model.
\begin{mythm}
 \label{theorem : main regret under strongly concave assumption}
 Under \cref{asp: lower bound of the cdf}, \cref{asp: linear combination of demand function}, \cref{asp: interior point}, \cref{asp: concavity}, \cref{asp:twice differentiable}, there exists a constant $\hat{C}$ such that the regret bound of our algorithm is
        $\mathcal{R}(T) \leq \hat{C}KT^{\frac{1}{2}}\log T.$
\end{mythm}

In \cref{decomposition of the regret}, the first term represents the regret during the exploration phase, which increases linearly with the total duration of the exploration. In the exploitation phase, we decompose the regret into two terms, where \(\hat{y}\) denotes the recommended inventory level and \(y_t\) denotes the actual implemented inventory level. This decomposition is based on the fact that \(y_t = \text{max}\{I_t, \hat{y}_t\}\).

Using Taylor expansion, the second term in \cref{decomposition of the regret} can be bounded as follows:
\begin{equation}
Q(x_t,p_t^*,y_t^*) - Q(x_t,{p}_t,\hat{y}_t) \leq  \hat{C}_2\|({p}_t,\hat{y}_t)-(p_t^*,y_t^*)\|_2^2.
\end{equation}

What remains is to bound the difference between the optimal price decision and the optimal inventory decision using the recommended strategy.
By utilizing the strong concavity of the function \(G\), we turn the difference of the price into the maximum of the differential of the function \(G\):
\begin{equation}
||p^*_t - {p}_t| \leq \frac{1}{\kappa_3} \left(\max_{p \in [p_{min}, p_{max}]}|\partial_pG - \partial_p\hat{G}|(x_t, p)\right).
\end{equation}

Next, we need to bound the difference between the first derivative of the true function \(G\) and the empirical estimate \(\hat{G}\), which can be bounded by the estimation error.

Similarly, the difference in inventory levels \(y_t^* - \hat{y}_t$ can be bounded using the definition and the bound of the price decision \(p^*_t - {p}_t\). Thus, we have the following theorem:

\begin{mythm} \label{theorem : the difference of the optimal decision and the recommended decision}
    In the exploitation phase, there exists a constant \(\tilde{C}\) such that
    \begin{equation}
    \begin{aligned}
    \mathbb{P}\Bigl(
    \|(p_t^*,y_t^*)-({p}_t,\hat{y}_t)\|_2
    &\leq \tilde{C}KT_0^{-\frac{1}{2}}(\log T_0)^{\frac{1}{2}}
    \Bigr) \\
    &> 1-18T_0^{-4}.
    \end{aligned}
    \end{equation}
\end{mythm}

Finally, we need to provide an upper bound for the third term in \cref{decomposition of the regret}.

In the case \(I_t \leq \hat{y}_t\), we have \(y_t = \hat{y}_t\), which results in zero regret. Below, we only consider the case where \(I_t > \hat{y}_t\):
\begin{align*}
    & |Q(x_t, {p}_t, \hat{y}_t) - Q(x_t, {p}_t, y_t)|\\
    =& |h\mathbb{E}[(\hat{y}_t-D_t)^+] + (b+{p}_t)\mathbb{E}[(D_t - \hat{y}_t)^+] \nonumber \\
    &\quad - h\mathbb{E}[(y_t-D_t)^+] -(b+{p}_t)\mathbb{E}[(D_t - y_t)^+]| \\
    \leq&  \max\{h, b+p_{max}\} (y_t - \hat{y}_t).
\end{align*}

Now we need to provide an upper bound on the difference \(y_t - \hat{y}_t\). To do this, we establish the connection between the contextual inventory level and the waiting time process. Unlike the method in \cite{joint_nonparametric}, which considers a non-contextual setting with the same optimal decision across all periods, we need to bound the difference in inventory levels \(y_t - \hat{y}_t\) dynamically.

To achieve this, we define the following two stochastic processes:
\begin{equation}
    A_{t+1} = [A_t - J_t]^+\quad \text{and} \quad B_{t+1} = [B_t + \frac{\rho_q}{\sqrt{t}} - J_t]^+   \label{def of stochastic process}
\end{equation}
where
\begin{equation}
    J_t =  \epsilon_t + \underline{\epsilon} - \overline{\epsilon} + \hat{\theta}^T\phi(x_{t+1}, p_{t+1}) -\lambda_{max}C_1K \frac{(\log T_0)^{\frac{1}{2}}}{T_0^{\frac{1}{2}}}.
\end{equation}
We have:
\[
y_t - \hat{y}_t \leq A_t \leq B_t
\]
for \(t \geq T_0 + 1\).

In the above stochastic processes, \(A_t\) serves as an intermediate process. Additionally, \(B_t\) can be interpreted as the waiting time of the $t$-th customer in the queuing system, where \(J_t\) is the interarrival time between the $t$th and $(t+1)$-th customer, and \(\rho_q/\sqrt{t }\) is the service time for the $t$th customer.

This idea is similar to the method presented in \cite{huh2009nonparametric}, which established a relationship between the amount of inventory exceeding the target level and the waiting time process in a GI/D/1 queue. However, they do not consider the impact of context and only need to bound the gradients to establish such a connection. In our analysis, since there are no gradients in the exploitation phase, we only need to exploit the estimation error during the exploration phase and the variability of features. Since this only involves the case where \(I_t > \hat{y}_t\), this method is also robust to perishable inventory levels.

\begin{myprop}\label{proposition: the upper bound of the true and recommended inventory}
    For any \(t \geq T_0 + 1\), \(y_t - \hat{y}_t \leq B_t\) with probability at least \(1 - \frac{2}{T_0^4}\), where \(B_t\) is defined in \cref{def of stochastic process}.
\end{myprop}

After providing a bound for the stochastic process defined above, we can upper bound the regret between the actual implemented inventory level and the recommended inventory level with the following proposition:

\begin{myprop}
\label{proposition : difference between the recommended inventory and the truly implemented inventory}
    There exists a constant \(C_4\) such that the regret from not achieving inventory targets during the commitment phase is bounded by
    $
    \mathbb{E}\left[\sum_{t=T_0+1}^T|Q(x_t, {p}_t,\hat{y}_t)-Q(x_t, {p}_t,y_t)|\right]\leq C_4 T^{\frac{1}{2}}
    $
    which holds with probability at least $1 - \frac{2}{T_0^4}.$
\end{myprop}

Summing up all the inequalities above, we can obtain the bound mentioned in \cref{theorem : main regret under strongly concave assumption}.

\begin{figure*}[t]
    \centering
    \includegraphics[width=\textwidth]{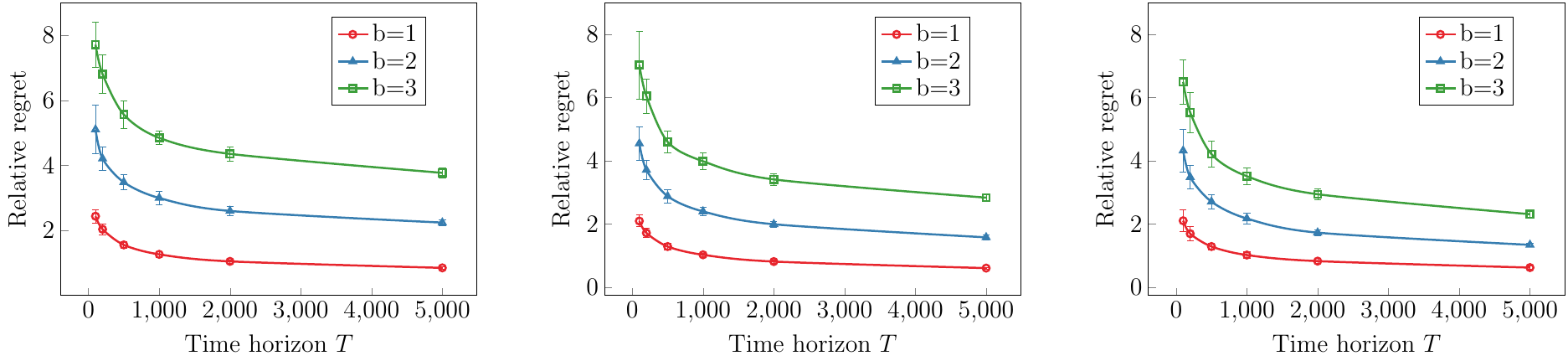}
    \caption{Relative regret over time horizon $T$ for different values of $h$ and $b$.}
    \label{fig: relative regret for different parameter values}
\end{figure*}

\begin{figure*}[t]
    \centering
    \includegraphics[width=0.96\textwidth]{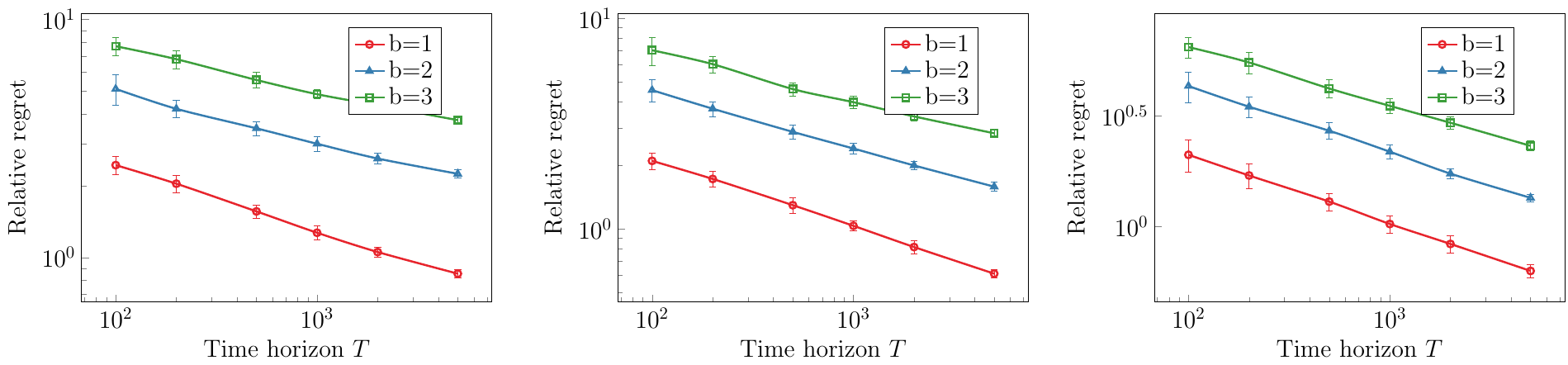}
    \caption{Relative regret over time horizon $T$ for different values of $h$ and $b$ (log-log scale).}
    \label{fig: relative regret for different parameter values in log-log scale}
\end{figure*}

\subsection{Non-concave Revenue Function}

The non-concave revenue function has been studied in \citet{joint_opt} in a non-contextual setting. They show that the penalty term related to inventory levels can destroy the concavity of the objective function, making the lost-sales model significantly more difficult to analyze compared to the backlogged case. To address this challenge, \citet{joint_opt} propose a search-based method for identifying the optimal price under the assumption that the revenue function is twice differentiable, with $\tilde{\Theta}(T^{3/5})$ minimax optimal regret bounds.

We extend their work in two key directions: we relax the assumption of second-order differentiability and generalize the problem to a contextual setting. In our case, the optimal prices vary over time due to changing contexts, and the revenue function \( G \) may lack smoothness, making the approach in \citet{joint_opt} inapplicable. Without assuming strong concavity of \( G \), we are unable to derive a precise upper bound on the difference between successive price decisions, as was done in the previous section. Instead, we adopt a greedy strategy to control this difference, which leads to a regret bound that is worse than the $\mathcal{O}(K\sqrt{T}\log T)$ rate achieved under more favorable assumptions.

\begin{mythm}
\label{theorem : regret of 2/3}
    With \cref{asp: differentiability of Q} and \cref{asp: linear combination of demand function},
    there exists a constant $\tilde{C}$ such that the regret bound of our algorithm is
    $\mathcal{R}(T) \leq \tilde{C}K^{\frac{2}{3}}T^{\frac{2}{3}}(\log T)^{\frac{1}{2}}.$
\end{mythm}

To prove \cref{theorem : regret of 2/3}, we decompose the regret into two parts:
\begin{enumerate}
    \item Bounding the revenue gap: \( G(x_t, p_t^*) - G(x_t, p_t) \)
    \item Bounding the inventory regret: \( Q(x_t, p_t, y_t^*) - Q(x_t, p_t, \hat{y}_t) \)
\end{enumerate}
For the first part, we use a decomposition of \( G(x_t, p_t^*) - G(x_t, p_t) \) and apply the estimation error bound for \( \hat{G}(x_t, p) \), leading to:
\begin{equation}
\begin{split}
G(x_t, p_t^*) - G(x_t, p_t) &\leq 2 \|G(x, p) - \hat{G}(x, p)\|_\infty \\
&\leq C \cdot KT_0^{-1/2}(\log T_0)^2.
\end{split}
\end{equation}
For the second part, we use an existing result in \cref{proposition : difference between the recommended inventory and the truly implemented inventory} to bound the inventory regret term.
Finally, combining these bounds results yields the desired results.

\begin{mythm}
\label{thm: lower bound}
    For any policy $\pi$, there exists a non-contextual problem instance whose revenue function is $m$-th differentiable, such that
    $
    \mathcal{R}(T) \geq \Omega(T^{(m+1)/(2m+1)})
    $
    for sufficiently large $T$.
\end{mythm}

The instance construction is motivated by \citep{Multimodal_DP}. The construction of the noise distribution starts with a smooth function \( u(x) \), which is normalized to define a function \( B(x) \), leading to a density function \( f(x) \) supported on \( [-1,1] \) with mean 0. Using this, a demand function \( \lambda_j(p) \) is defined, which perturbs the base revenue function \( r_0(p) \) using a small parameter \( \eta \) and a bump function. The reward function \( Q_j(p, y) \) is shown to be $m$-th differentiable, with bounds on the Hellinger distance between different environments. A lower bound on expected regret is derived using the pigeonhole principle, showing that for some interval \( i^* \), the regret for any policy is at least \( \Omega(T^{\frac{m+1}{2m+1}}) \).

Combing the above upper and lower bounds, we conclude that $\Theta(T^{\frac{2}{3}})$ is an achievable minimax optimal regret bounds only under the differentiabilty assumption.

%% file: sessions/numerical_experiments.tex
\section{Numerical Experiments}

We conduct a numerical study to evaluate the empirical performance of our algorithms. 

\subsection{Test on parameters $h$ and $b$}
The experiments are based on a linear demand model:  
\[
d_t = \alpha + x_t^\top \beta + \epsilon_t,
\]  
where the parameter vector \( (\alpha, \beta) \in \mathbb{R}^4 \) is sampled from a standard normal distribution and normalized to unit norm. The noise term \( \epsilon_t \) is drawn independently from the uniform distribution over \([-1, 1]\). Prices are constrained to the range \( p_{\min} = 0.1 \) and \( p_{\max} = 2 \).

For each experimental setting, we perform 50 independent runs. At each round, context vectors \( x_t \) are drawn from a standard normal distribution and normalized to unit length. We evaluate performance using relative regret, defined as \( \frac{\mathcal{R}(T)}{T} \), where \( \mathcal{R}(T) \) is the cumulative regret up to time \( T \). Results are reported for time horizons \( T \in \{100, 200, 500, 1000, 2000, 5000\} \), including the mean and standard deviation across trials.

The numerical experiments in \cref{fig: relative regret for different parameter values} show that relative regret decreases as the time horizon \( T \) increases. This trend holds across all three subplots with different parameter settings, indicating that the algorithm improves over time by leveraging more data for better decision-making and a balanced exploration-exploitation trade-off.

The error bars, representing standard deviation across trials, highlight increased variability at smaller \( T \), due to insufficient data for reliable convergence. As \( T \) grows, the decreasing variability reflects improved stability. In \cref{fig: relative regret for different parameter values in log-log scale}, the log-log plot shows a linear trend, suggesting a power-law relationship between regret and \( T \), which supports the algorithm's robustness across different parameter settings.

\subsection{Test on dimension $K$}

We investigate how the algorithm's performance depends on the dimension of the basis functions, denoted by $K$. We vary $K$ from $4$ to $12$ to assess its impact on regret.
To ensure a fair comparison across different values of $K$, we fix the total time horizon at $T = 1000$, and set the cost parameters as $h = 1$ and $b = 1$. All other experimental settings remain consistent with the base configuration described earlier.

\cref{fig: relative regret for different K values} shows the relative regret across different $K$ values. In summary, our results highlight key insights into the trade-offs involved in contextual joint pricing and inventory control:

\begin{figure}[h]
    \centering
    \includegraphics[width=0.45\textwidth]{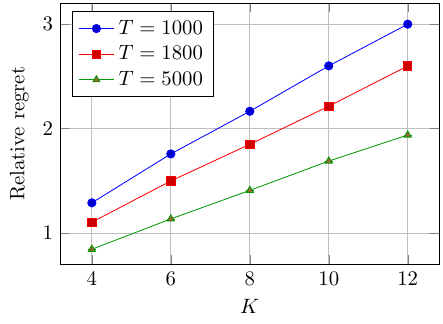}
    \caption{Relative regret across different $K$ values with standard deviation error bars.}
    \label{fig: relative regret for different K values}
\end{figure}

\begin{itemize}
    \item \textbf{Regret Increases with Model Complexity:} The relative regret increases consistently with $K$, indicating that higher-dimensional models are more difficult to learn under a fixed time horizon. Intuitively, increasing $K$ enhances the expressive power of the model but requires more exploration.

    \item \textbf{Stable Performance Across Trials:} Despite the increased complexity, the algorithm maintains controlled variance, with standard deviations remaining moderate across different values of $K$, indicating robust and reliable performance.

\end{itemize}

%% file: sessions/conclusion.tex
\section{Conclusion}
This paper studies the joint pricing and inventory control problem under a linear additive model with contextual information and censored demand observations. We address the critical gap where existing approaches fail to account for both contextual factors and the difficulty of lost sales that obscure true demand information.

Our main contributions include: (1) a context-aware framework for joint pricing and inventory management where both pricing and inventory policies adapt to changing contexts; (2) an easy-to-implement algorithm that achieves optimal regret bounds of $\mathcal{O}(K\sqrt{T}\log T)$ under concave revenue conditions and $\mathcal{O}(K^{2/3}T^{2/3}(\log T)^{1/2})$ for the general case, with matching lower bounds confirming minimax optimality; and (3) a queueing-theoretic method to handle infeasible target inventory levels, ensuring stable performance under stochasticity. Extensive numerical experiments demonstrate the algorithm's effectiveness across diverse scenarios.

For future work, it would be valuable to investigate this problem under alternative modeling assumptions beyond the linear additive framework. It is also of interest to develop UCB-style analyses tailored to censored demand, which could offer principled exploration under partial observability.